%% file: main-ICLR.tex
\documentclass{article} 
\usepackage{iclr2021_conference,times}

\usepackage{amsmath,amssymb,amsfonts}
\usepackage{algorithmic}
\usepackage{graphicx}
\usepackage{microtype}
\usepackage{graphicx}
\usepackage{subfigure}
\usepackage{booktabs} 
\usepackage{textcomp}
\usepackage{xcolor}

\usepackage{booktabs}
\usepackage{multirow}
\usepackage{siunitx}
\usepackage{makecell}

\usepackage{amsmath}
\usepackage{float}
\usepackage{color}
\usepackage{xcolor}
\usepackage{hyperref}
\usepackage[linesnumbered,ruled]{algorithm2e}
\usepackage{subfigure}
\usepackage{booktabs}
\usepackage{siunitx}

\usepackage[utf8]{inputenc} 
\usepackage[T1]{fontenc}    
\usepackage{url}            
\usepackage{booktabs}       
\usepackage{amsfonts}       
\usepackage{nicefrac}       
\usepackage{microtype}      

\usepackage{amsthm,amssymb,amsmath,mathtools}

\title{Personalized Federated Learning by Structured and Unstructured Pruning under Data Heterogeneity}

\author{
  Saeed Vahidian 
  \thanks{
  Equal contribution. Correspondence to Saeed@ucsd.edu}
  \\
  \And
  Mahdi Morafah\footnotemark[1]\\
  \And
  Bill Lin\\
  \AND
  \textnormal{\small{ Dept. of Electrical and Computer Engineering}} \\
  \small{~University of California San Diego} \\
}

\usepackage{xcolor}
\usepackage{etoolbox}
\usepackage{ulem}

\providebool{comments}
\booltrue{comments}
\ifbool{comments}{
\newcommand{\BL}[1]{{\color{olive} {\bf Lotzi:} #1}}
\newcommand{\SK}[1]{{\color{blue} {\bf Siavash:} #1}}

\newcommand{\SKremovesuggestion}[1]{\color{red} \sout{#1}}

\newcommand{\SZ}[1]{{\color{violet} #1}}
}{
\newcommand{\BL}[1]{}
\newcommand{\SK}[1]{}

\newcommand{\SKremovesuggestion}[1]{}
\newcommand{\SZ}[1]{}
}

\iclrfinalcopy
\begin{document}

\maketitle

\begin{abstract}
The traditional approach in FL tries to learn a single global model collaboratively with the help of many clients under the orchestration of a central server. However, learning a single global model might not work well for all clients participating in the FL under data heterogeneity. Therefore, the personalization of the global model becomes crucial in handling the challenges that arise with statistical heterogeneity and the non-IID distribution of data. Unlike prior works, in this work we propose a new approach for obtaining a personalized model from a client-level objective. This further motivates all clients to participate in federation even under statistical heterogeneity in order to improve their performance, instead of merely being a source of data and model training for the central server. To realize this personalization, we leverage finding a small subnetwork for each client by applying hybrid pruning (combination of structured and unstructured pruning), and unstructured pruning. Through a range of experiments on different benchmarks, we observed that the clients with similar data (labels) share similar personal parameters. By finding a subnetwork for each client rather than taking the average over all parameters of all clients for the entire federation as in traditional FL, we efficiently calculate the averaging on the remaining parameters of each subnetwork of each client. We call this novel parameter averaging as Sub-FedAvg. Furthermore, in our proposed approach, the clients are not required to have knowledge of any underlying data distributions or label similarities among the rest of clients. The non-IID nature of each client’s local data provides distinguishing subnetworks without sharing any data. We evaluate our method on federated image classification with real world datasets. Our method outperforms existing state-of-the-art.

\end{abstract}

\input{sections/Introduction}

\input{sections/RelatedWork}
\input{sections/Method}
\input{sections/Experiments}
\input{sections/Conclusion}


\bibliography{Saeed}

\bibliographystyle{iclr2021_conference}

\end{document}

%% file: sections/Introduction.tex
\section{Introduction}

Federated learning (FL) refers to the paradigm of learning a common objective collaboratively with the help of many clients (e.g. mobile devices or data centers) under coordination of a central server. Such decentralized paradigms has recently drawn significant attention in the context of machine learning and deep learning as they provide several advantages such as scalability to larger datasets, data locality, ownership and privacy, compared to traditional, centralized learning approaches~(\cite{mcmahan2017communication}). In this context, a trusted server aggregates parameters optimized in a decentralized fashion by multiple clients. The resulting model is then distributed back to all clients, ultimately converging to a joint representative model without explicitly having to share the data~(\cite{mcmahan2017federated}). 

Google has pioneered cross-device FL where the emphasis is on edge device applications~(\cite{hard2018federated, chen2019closer}). For instance, Google makes use of FL in the Gboard mobile keyboards, in features on Pixel phones~(\cite{hard2018federated}), and in Android Messages~(\cite{android-messages}). Now, cross-device FL and federated data analysis are being widely applied in electronic devices, such as cross-device FL in iOS 13, and ``Hey Siri”~(\cite{apple-siri}), etc. Nonetheless, much research has been done recently to address challenges associated with FL, including statistical challenges, the communication cost of sending large scale matrices of parameters of deep networks~(\cite{konevcny2016federated}), computing constraints, and personalization~(\cite{yu2020salvaging}). 

FL faces statistical heterogeneity due to fact that the distribution of the data across the clients is inherently non-IID~(\cite{zhao2018-non-iid}). In practice, it is an unrealistic assumption that the local data on each client is always IID~(\cite{mcmahan2017communication}). The original goal of FL, training a single global model on the union of client datasets, becomes harder with non-IID data~(\cite{chen2018federated-fedmeta}). To address the non-IID challenge,~(\cite{mcmahan2017communication})
demonstrated that FedAvg can work with certain non-IID data. Other studies put efforts to alleviate the statistical heterogeneity via performing personalization in FL~(\cite{jiang2019improving,  wang2019federated,mansour2020three}). Communication cost as another major bottleneck of FL were studied in the literature. In particular,~(\cite{bonawitz2017practical}) proposed an aggregation method for FL, allowing a central server to perform computation of high-dimensional data from mobile devices.~(\cite{konevcny2016federated}) suggested structured and sketched updates which reduce communication cost by two orders of magnitude. 

\subsection{Federated Learning}
In this work, our focus is on the non-IID properties of the clients’ data, as well as the critical factor of the communication cost and personalizations. A FL scenario must also take care of a myriad of practical issues including clients' datasets that change as data is added and deleted; availability of the clients, corrupted updates by the clients, etc. Studying all of these challenges are beyond the scope of the present work; instead, we propose a framework for FL to address the three key issues of statistical heterogeneity, communications cost, and personalizations.

\subsection{Contributions}
In our work, we argue that the mentioned three challenges point in the same
direction, i.e., we show that a single solution can be proposed to address all the challenges simultaneously. Our main contributions are:

\begin{itemize}
\item In the current work, we consider a realistic scenario, where each client owns limited data with non-IID settings. Here we focus on FL from a client-level or personalized perspective. Motivated by the fact that, in practice, the participation of clients in federation is contingent upon satisfying their objectives, we leverage the statistical heterogeneity as a blessing factor and propose a new framework for personalized federated learning. In the proposed method, the clients are not only a source of data and model training for the global serve, but also can ameliorate the performance on their personalized target distributions.

\item Other than that, our underlying method is a straightforward yet effective solution that further alleviates the communication bottleneck as well in FL by finding a subnetwork for each client by a novel hybrid (combination of structured and unstructured) and unstructured pruning strategy on the neural network models of the clients. To the best of our knowledge, there has not been any proposed method in FL to address the mentioned critical challenges simultaneously. 

\end{itemize}

%% file: sections/RelatedWork.tex
\section{Related Work}
In FL, the edge devices carry most of the load of computation and a central server updates the model parameters using the descending directions returned by the edge devices. However, FL has three unique characteristics that makes it different from the parallel optimization systems in the following aspects: \\

\noindent \textbf{Statistical Heterogeneity}
The main motivation for clients to take part in FL is to improve their own model performance. Especially, the clients who have very limited private data benefits the most from collaboratively learned models. However, for some of the clients who have enough private data, there is not much benefit to participate in FL. This issue becomes even worse in case of statistical heterogeneity of the clients. In fact, due to the non-IID distribution of data across devices, it leads to scenarios where some participants may gain no benefit by participating in FL since the global shared model is less accurate than the local models that they can train on their own (~\cite{yu2020salvaging, Hanzely2020}).\\

\noindent \textbf{Communication Efficiency}
A naive implementation of FL framework entails each client sends a full
model update back to the central server in each communication round. For large neural networks, this step will be the bottleneck of FL due to the asymmetric nature of internet connection speeds: the uplink is typically much slower than the downlink. The US average internet speed was 55Mbps download vs. 18.9Mbps upload, and with some internet service providers, this issue is worse, e.g., Xfinity provides 125Mbps down vs. 15Mbps up~(\cite{ speedtest}). Hence, it is important to propose methods that necessitate less uplink communication cost~(\cite{mcmahan2017communication, Bonawitz2017, han2018geometric, songdeep2017, LG2020, reisizadeh2019robust}). For instance, some existing model can reduce the communication cost with structured updates, and sketched updates~(\cite{konevcny2016federated}). Others do so by compressing the gradients~(\cite{songdeep2017}). \\

\noindent \textbf{Personalizations and Accuracy for Individual Clients}
FL is explicitly designed for non-IID clients, but most of prior works measured global accuracy and not accuracy for individual clients. A global model can perform well on personalized predictions if the client’s context and personal data is nicely featurized and embodied in the dataset, which is not the case in most clients~(\cite{Kairouz2019}). Most techniques for personalization either affect privacy or involve two separate steps where a global model is constituted collaboratively in the first step, and then the global model is personalized for each client using the client’s private data in the second step. These two steps might add extra computational overhead~(\cite{jiang2019improving,  yu2020salvaging,sim2019investigation}).

%% file: sections/Method.tex
\section{Method}

Deep neural networks (DNN), especially deep Convolutional Neural Networks (CNN), have achieved significant success in various tasks and applications. However, the excellent performance of modern CNNs comes often at significant inference costs due to more stacked layers, and thus more learnable parameters. The usage of these high capacity networks may be largely hindered for FL scenarios where in addition to accuracy, computational efficiency and small network sizes are crucial enabling factors. For example, a ResNet-152 has more than $60$ million parameters and entails more than $20G$ float-point-operations (FLOPs) when training with an image with resolution $224 \times
224$~(\cite{he2016deep}). This is unlikely to be affordable on resource constrained platforms such as embedded edge devices~(\cite{li2016pruning}). 

The recent strategies of ameliorating CNN efficiency mostly focus on compressing models and accelerating inference without significantly sacrificing their accuracy performance. Among adopted methods, progressive pruning appears to be an outstanding one where a deep neural net is trained, then pruned, and then fine tuned to restore performance~(\cite{franklelottery2018}).


\subsection{ Efficient Learning with Pruning}
In this paper, we show that in a FL scenario, there consistently exist smaller similar subnetworks for clients with similar labels of data that can improve accuracy of each other in FL. Based on these results, we state the following observation.\\
\noindent\textbf{Client Subnetwork Observation:} \textit{Through extensive experiments on various DNNs and benchmark datasets, we noticed the existence
of similar subnetworks for clients with even partially similar data (labels) and the cheap costs needed to reliably find them, and develop a new algorithm to improve their accuracies.}

More specifically, consider a dense feed-forward neural network model $f(x; \theta)$ with initial parameters $\theta = \theta_0 \sim D_{\theta}$ for each client as well as the central server.  Further assume that optimizing client $k$’s neural net with stochastic gradient descent (SGD) on its own training set, $f(x; \theta_k)$, reaches minimum validation loss $L_k$ at iteration $ j$ with test accuracy $Acc_{k}$. Besides, consider training $f(x; m_k \odot \theta_k)$ with a mask $m \in \{0, 1\}^{|\theta|}$ on client $k$'s parameters reaches minimum validation loss $L'_k$ when being optimized with SGD on the same training set at the $j'$ iteration of training with test accuracy of $Acc_k'$. We observed that under some scheduling $Acc_k \le Acc_k'$ which means the improvement of the accuracy of each client $k$ while the $p$-percentage of each client's network is pruned. The following sections delineate the conceptual overview of accuracy improvement in our proposed FL method.


\subsection{Efficient Learning with Pruning}

Consider a dense feed-forward neural network model $f(x; \theta)$ with initial parameters $\theta = \theta_0 \sim D_{\theta}$ for each client as well as the central server.  Further assume that optimizing client $k$’s neural net with stochastic gradient descent (SGD) on its own training set, $f(x; \theta_k)$, reaches minimum validation loss $L_k$ at iteration $ j$ with test accuracy $Acc_{k}$. Besides, consider training $f(x; m_k \odot \theta_k)$ with a mask $m \in \{0, 1\}^{|\theta|}$ on client $k$'s parameters reaches minimum validation loss $L'_k$ when being optimized with SGD on the same training set at the $j'$ iteration of training with test accuracy of $Acc_k'$. We observed that under some scheduling $Acc_k \le Acc_k'$ which means the improvement of the accuracy of each client $k$ while the $p$-percentage of each client's network is pruned. The following sections delineate the conceptual overview of accuracy improvement in our proposed FL method.


\subsection{Structured, Unstructured, and Hybrid Pruning}
In this paper, we aim to find a smaller subnetwork for each client in the FL scenario via three different pruning levels, i.e., channel level pruning (structured), parameter level pruning (unstructured), and hybrid (combination of structured and unstructured) level pruning in deep CNNs. Unstructured-level pruning is motivated by the fact that it gives the highest flexibility and generality for compression rate~(\cite{han2015deep}
)for both shallow and deep neural networks. On the other hand, channel level pruning is less flexible as some whole layers need to be pruned. Pruning a whole layer is only effective when the neural networks is sufficiently deep~(\cite{wen2016learning}). Nonetheless, channel-level pruning provides a fair trade-off between flexibility and ease of implementation. There is also a hybrid level pruning which is realized by the combination of channel and parameter level pruning. It can be applied to any typical CNNs or fully connected networks, which results in a compressed network that can be efficiently inferenced on conventional CNN platforms. 


\subsection{Why Learning with Pruning in FL is Efficient}

We consider synchronized algorithms for FL, and we suggest a round to consist of the following steps:
$i)$ A subset of existing clients is randomly selected, each of which downloads the current model parameters from the server.
$ii)$ Each client in the subset computes an updated model based on its local data and prunes its neural network according to Algorithm 1 or 2 to obtain its own subnetwork.
$iii)$ The model updates are sent from the selected clients to the sever.
$iv)$ The server aggregates these models by applying the Sub-FedAvg method where the average is taken only on the intersection of the remaining channels (in structured pruning) or the remaining parameters (in unstructured pruning) of each client to construct an improved global model. 

In what follows, we will first provide a conceptual overview of this method of pruning and the proposed Sub-FedAvg on the server by the following remark and then delineate the routine of the algorithm, and finally show the evaluation performance by benchmarking it with state-of-the-art methods of FL in DNN models based on representative datasets.\\

\noindent \textbf{Remark-1:} At the beginning of each communication round, each client downloads the model from the server which contains the featured and embodied data of all clients and starts training on its local data. Due to the statistical heterogeneity of clients, part of the channels (filters) and parameters are personalized to each client. By iteratively pruning the parameters and channels, we remove the commonly shared parameters of
each layer and keep the personalized parameters that can represent the features of local data in each client. Since the clients with similar data (label overlap) have share similar personalized parameters, by the proposed Sub-FedAvg we would average the models of clients on the server on the intersection of remaining channels and parameters of the clients. This method of aggregation on the server not only does not impact the accuracy performance of each client in a non-IID setting but also improves it significantly as reported in Table~\ref{tabel-1}.


\subsection{Algorithm}
\noindent \textbf{Unstructured Pruning} At each communication round, the server randomly samples a set of clients $\mathcal{S}$ and sends the model to the selected clients. Each client $C_k$ starts training the local model. Given a target pruning ratio $p$ and a pruning percentage, $r_{us}$, for each communication round, a binary mask is derived at the end of the first epoch, and at the end of last epoch w.r.t. the full dense network of client $C_k$ where assigns 0 to the lowest $r_{us}$-percent of the absolute value of parameters and $1$ to the rest. 
To this end, after training, each client tests its model on the validation data $D^{val}_k$. If the validation accuracy is above a pre-considered threshold $Acc_{th}$ and if the target pruning rate is not achieved yet and finally, if the Hamming distance between the two masks (mask distance) is above a pre-defined threshold $\epsilon$, the client $C_k$ prunes its model with the mask obtained at the end of the last epoch. This process is iterated in all communication rounds till the conditions are not satisfied. \\

\noindent \textbf{Structured Pruning} We adopt the same channel pruning as in~(\cite{liu2017learning}) since it is hardware friendly and aligns best with our mission of efficient training and performance improvement. Herein, we do the same process as in unstructured pruning and follow~(\cite{liu2017learning}) to consider the scaling factor $r$ in batch normalization (BN) layers as indicators of the corresponding importance of channels. We derive a mask for each filter individually and channels of filters are pruned based on pruning threshold. For simplicity, the pruning threshold is determined by a percentile among all scaling factors, e.g., $p\%$ of channels are pruned.\\

\noindent \textbf{Hybrid Pruning}
Structured pruning is more effective when the depth of the neural network of clients are sufficiently large~(\cite{huang2016deep}). On the other hand, due to the benefit of structured pruning in enabling the model with structured sparsity and more efficient memory usage and more computation acceleration, we would get the benefit of it even in small network by leveraging the combination of structured and unstructured pruning, namely hybrid pruning. In this method, we apply the structured pruning on the filters and employ unstructured pruning on the fully connected layers. Algorithm~1 and~2 summarize our methodology. \textit{It is noteworthy that in Algorithm~2, the structured and unstructured pruning process are independent of each other meaning that when one does satisfy the constraints it applies the mask regardless of if the other one satisfies the constraints or not.} Moreover, the way that the parameters are aggregated on the server is called Sub-FedAvg in this paper.  

\begin{algorithm}[t]
\label{alg1}
\caption{Sub-FedAvg with unstructured pruning (Sub-FedAvg (Un))}
\SetKwInOut{Require}{require}
\textbf{Server:} initialize the server model with $\theta_g$.\\
\Require{$k \leftarrow {\rm{max}}(K\times N)$: $N$ number of available clients, sampling rate K}

$\mathcal{S}_t \leftarrow$ (random set of $k$ clients)
\For {each round j = 1, 2,...} {
\For {each client $k \in {{\mathcal{S}}_j}$ \rm{\textbf{in parallel}}}
    {
     download $\theta^{j}_g $ from the server and start training; set the pruning ratio $r_{us}$ and target pruning rate $p_{us}$\\
     
     derive the mask $ m^{j, fe}_k$ from $\theta^{j, fe}_k$ at the end of first epoch (fe)\\
    derive the mask $m^{j, le}_k$ from $\theta^{j, le}_k$ at the end of last epoch (le)\\

    $\theta^{j+1}_k\leftarrow {\rm{ClientUpdate}}(C_k; \theta^j_k ; \theta_0)$: using SGD training\\
    }
    
    $\theta^{j+1}_g\leftarrow$ aggregate subnetworks of clients, ~${\theta^{j+1}_k}$, and take the avg on the intersection of unpruned parameters (Sub-FedAvg).
}
\textbf{{ClientUpdate}}$(C_k; \theta^j_k ; \theta_0)$:\\
evaluate $\theta^{j}_k$ on the local validation data $D^{val}_k$ and report the accuracy.\\
$\Delta_{us}\leftarrow$ Mask Distance of $|m^{j, fe}_k-m^{j, le}_k|$\\
\If{\rm{accuracy}  $\ge Acc_{th}$ $\&$ 
\rm{target pruning rate} $p$ is not achieved yet $\&$ $\Delta_{us} \ge \epsilon_{us}$ }
{ 
$\theta^{j+1}_k=\theta^{j, le}_k \odot m^{j, le}_k$ : apply the mask

}
\Return $\theta^{j+1}_k$ to server
\end{algorithm}


\begin{algorithm}[t]
\label{alg2}
\caption{Sub-FedAvg with hybrid pruning (Sub-FedAvg (Hy))}
\SetKwInOut{Require}{require}
\textbf{Server:} initialize the server $(\theta_g)$ and clients $(\theta_0)$.\\
\textbf{Clients:} set the scaling factor $r$, the pruning ratios $r_{us}$ and $r_{s}$, the target pruning rates $p_{us}$, and $p_{s}$; the mask distances $ \epsilon_s$, and $ \epsilon_{us}$.\\
\Require{$k \leftarrow {\rm{max}}(K\times N)$: $N$ number of available clients, sampling rate K}

$\mathcal{S}_t \leftarrow$ (random set of $k$ clients)

\For {each round j = 1, 2,...} {
\For {each client $k \in {{\mathcal{S}}_j}$ \rm{\textbf{in parallel}}}
    {
     download $\theta^{j}_g $ from the server and start training.\\
     
     derive the mask by structured pruning based on $r_{s}$ towards the target ratio $p_{s}$, at the end of first epoch ($m_k^{(1)}$) and last epoch ($m_k^{(2)}$).\\

      derive the mask $ m^{j, fe}_k$ from $\theta^{j, fe}_k$ by unstructured pruning on the fc-layers at the end of first epoch (fe) \\
      and derive the mask $m^{j, le}_k$ from $\theta^{j, le}_k$ by unstructured pruning on the fc-layers at the end of last epoch (le)\\
    
    $\theta^{j+1}_k\leftarrow {\rm{ClientUpdate}}(C_k; \theta^j_k ; \theta_0)$: using SGD training.\\
    }
 
    $\theta^{j+1}_g\leftarrow$ aggregate subnetworks of clients~${\theta^{j+1}_k}$
}
\textbf{{ClientUpdate}}$(C_k; \theta^j_k ; \theta_0)$:\\
evaluate $\theta^{j}_k$ on the local validation data $D^{val}_k$ and report the accuracy.\\
$\Delta_{s}\leftarrow$ Mask Distance $|m^{(1)}_k-m^{(2)}_k| $\\
$\Delta_{us}\leftarrow$ Mask Distance of $|m^{j, fe}_k-m^{j, le}_k|$

\If{ \rm{accuracy}  $\ge Acc_{th}$ $\&$ target pruning rate $p_{s}$, $p_{us}$ are not achieved yet:}
{\If{ \rm{any of the conditions} $\Delta_{s} \ge \epsilon$ or $\Delta_{us} \ge \epsilon$ \rm{hold: }}
{  \rm{apply its corresponding mask}\\
$\theta^{j+1}_k=\theta^{j, le}_k \odot m^{(2)}_k\odot m^{j, le}_k$}
}

\Return $\theta^{j+1}_k$ to server

\end{algorithm}

%% file: sections/Experiments.tex
\section{Experiments}

In this section, we provide extensive experimental results evaluating our proposed algorithms. The code of this work is publicly available at: \url{https://github.com/MMorafah/Sub-FedAvg}. 
\subsection{Experiment Settings}
\noindent \textbf{Datasets and Non-IID Partitions} We use MNIST (\cite{lecun2010mnist}), CIFAR-10 (\cite{Krizhevsky09learningmultiple}), EMNIST (\cite{cohen_afshar_tapson_schaik_2017}), and CIFAR-100 datasets in our experiments. To produce non-IID partitions, we partition all the training dataset into shards of 250 examples (except for CIFAR-100 where we use 125 examples) and randomly assign two shards to each client. Evaluation data for each client is all the test set for the training dataset labels they have.\\

\noindent \textbf{Architecture} The architecture we used for MNIST, and EMNIST datasets is a $5$-layer CNN consisting of two layers of $5\times5$ convolutional layer with $10$ and $20$ channels respectively, and each convolutional layer is followed by batch-normalization and $2\times2$ max pooling layers. Finally, a fully-connected layer with 50 units followed by another fully-connected layer with 10 units to produce the logits ($30900$ total parameters + $30$ channels). For all layers, except the last layer, we use the ReLU non-linear activation function.  
For CIFAR-10 and CIFAR-100 datasets we use LeNet-5 (62000 total parameters + 16 channels) architecture (\cite{lecun1998gradient}). In the LeNet-5 architecture, we also add a batch-normalization layer after each convolutional layer.\\

\noindent \textbf{Hyper-parameter Setting} For all experiments, we setup 100 clients with local batch size $10$, local epoch $5$, and an SGD optimizer with learning rate and momentum of $0.01$ and $0.5$, respectively. Further, the threshold for distance of masks are $10^{-4}$, and $0.05$ for unstructured and, hybrid pruning algorithms, respectively. 


\begin{figure*}[!h]
    \centering
    \begin{tabular}{ccc}
          \includegraphics[width=.25\pdfpagewidth]{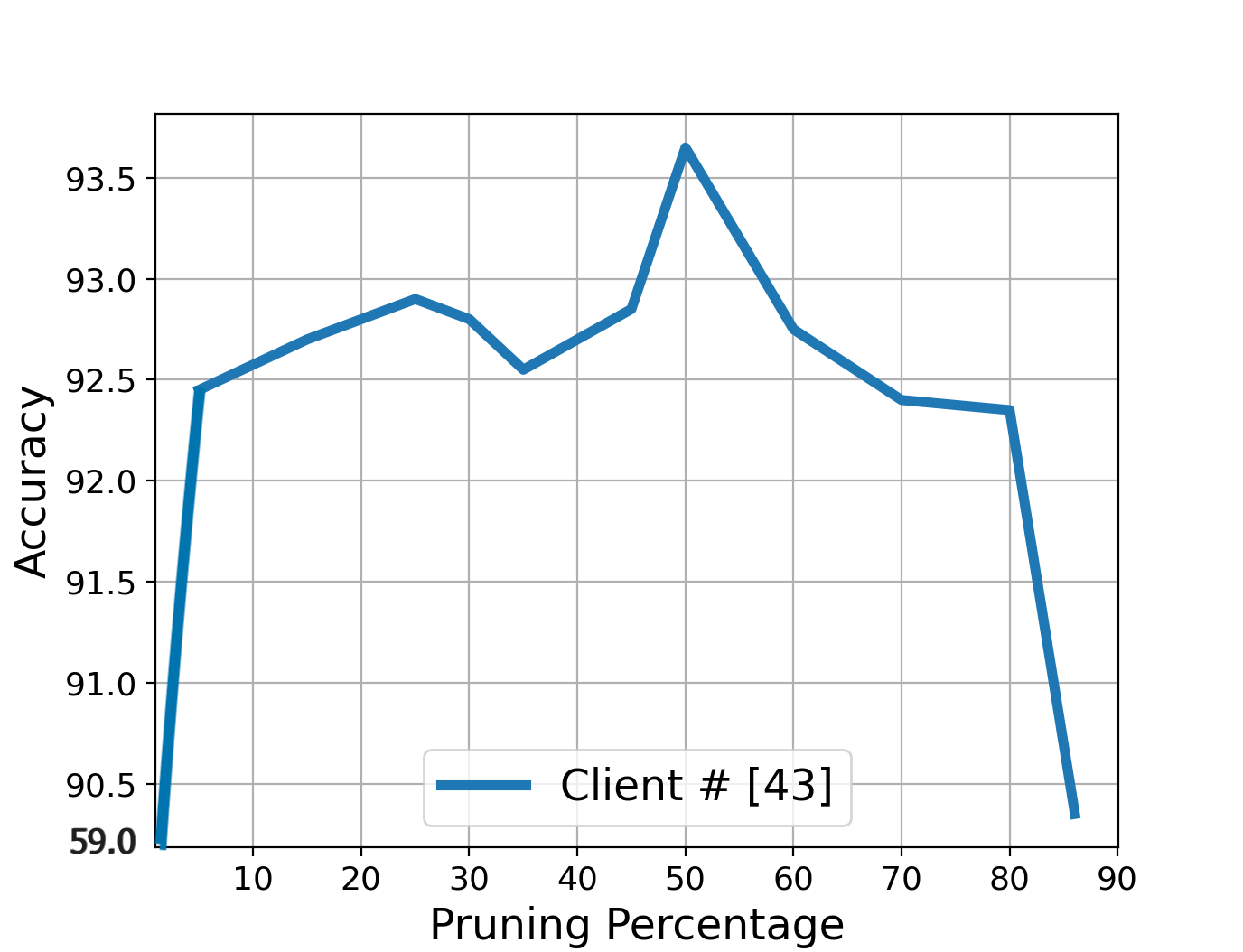} \hspace{-8mm} & 
          \includegraphics[width=.24\pdfpagewidth]{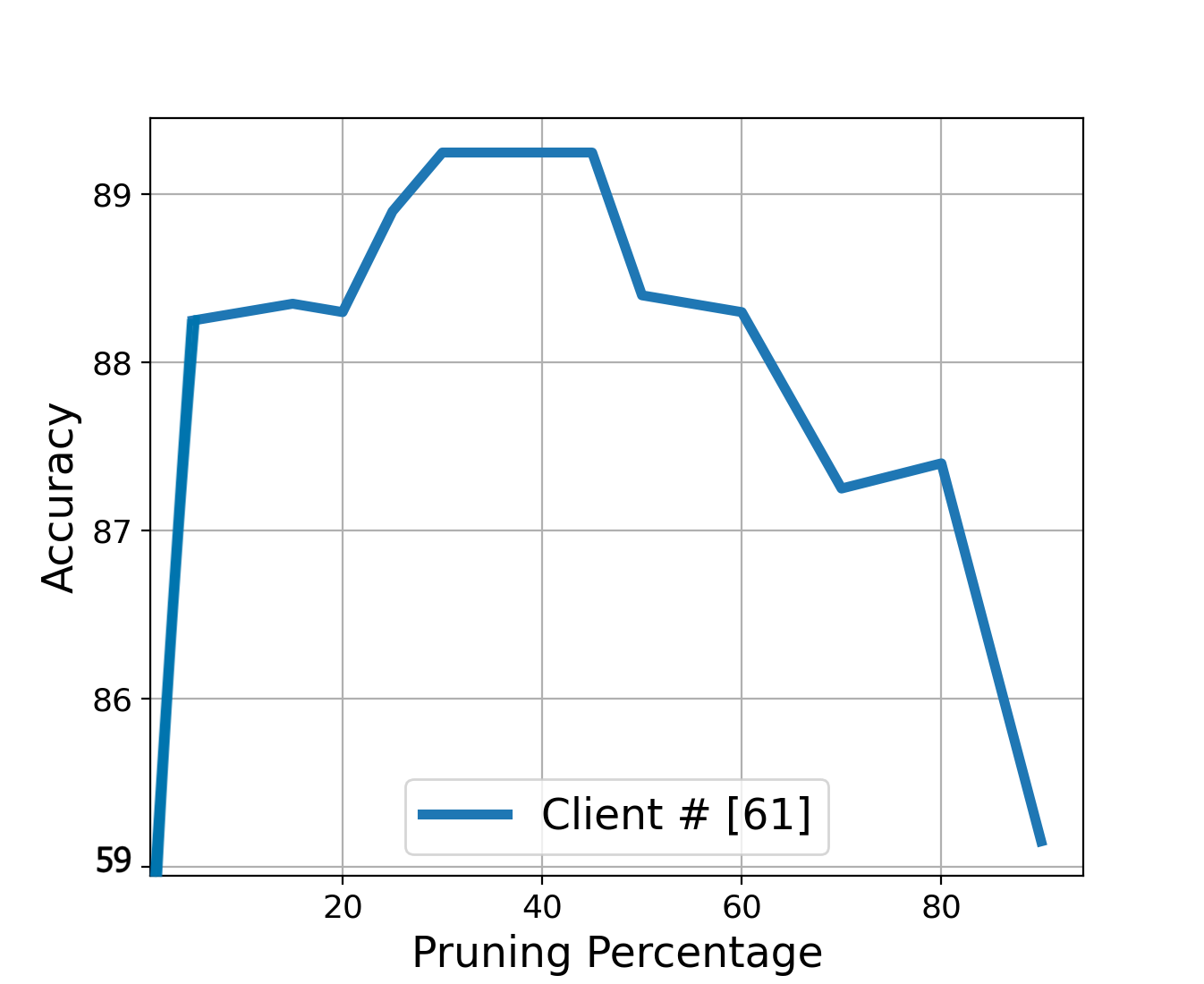} \hspace{-8mm}
          &
          \includegraphics[width=.24\pdfpagewidth]{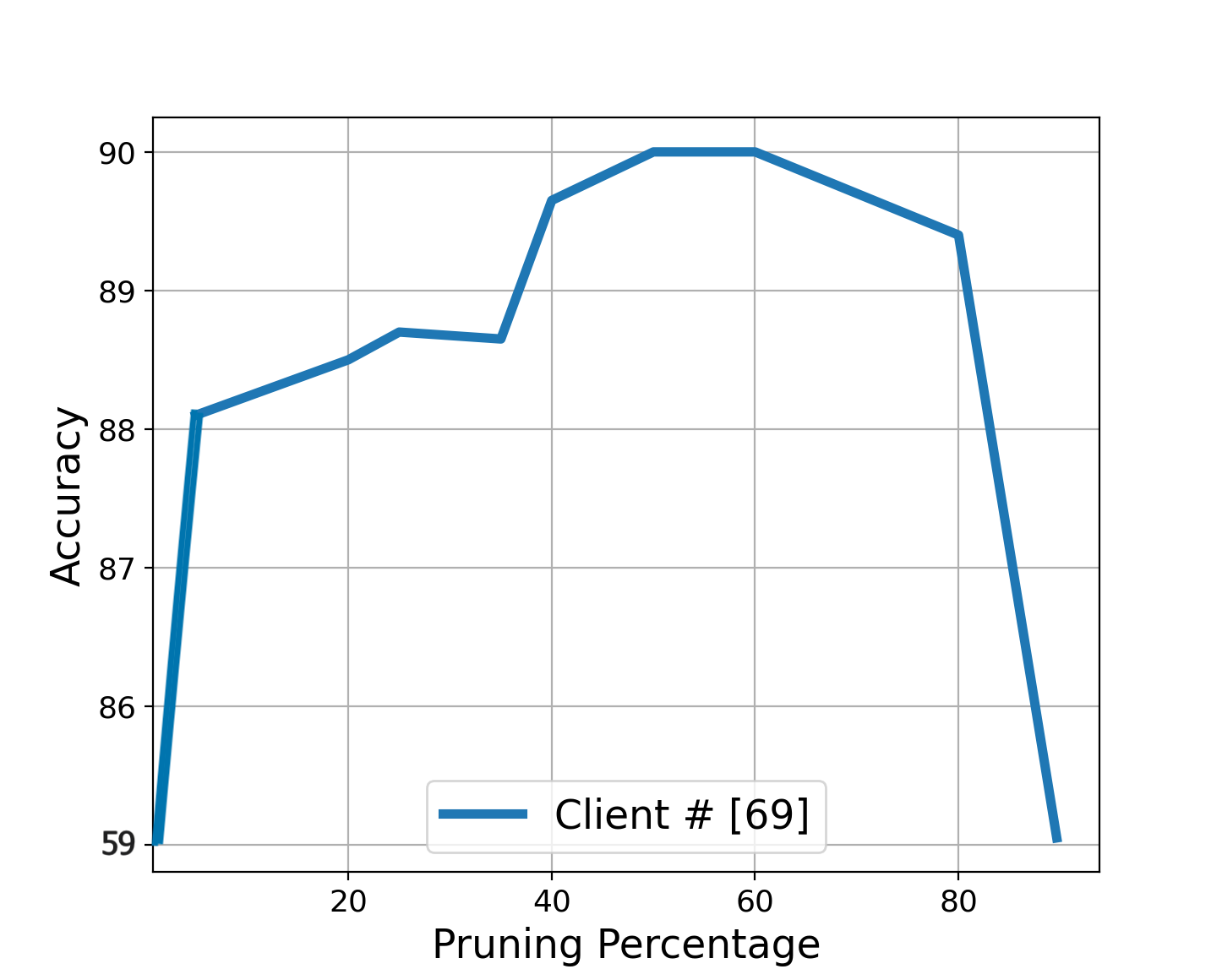}
    \end{tabular}
    \caption{Test accuracy vs pruning percentage result of the proposed Sub-FedAvg (Un) on some sampled clients on LeNet-5 for CIFAR-10. }\label{fig-subnet}
\end{figure*}


\subsection{Main Results}
We train LeNet-5 architecture on CIFAR-10/100, MNIST, and EMNIST datasets in a non-IID data distribution setting. We compare the results of our proposed algorithms against the state-of-the-art baselines. Table~\ref{tabel-1} reports the results for the average accuracy across all clients, communication cost, and the FLOP and parameters reduction. We first focus on the accuracy comparison against the baselines. The accuracy is measured for both cases when the subnetworks are drawn by unstructured pruning, and hybrid pruning. For unstructured pruning, we evaluate the accuracy when $30\%$, $50\%$, and $70\%$ of the parameters are pruned. This is while for the hybrid pruning we report the accuracy results when $50\%$, $70\%$, and $90\%$ of the parameters are pruned.

We can clearly see the advantage of the proposed Sub-FedAvg (Hy) and Sub-FedAvg (Un) algorithms over the state-of-the-arts in improving the accuracy performance of clients. This is realized due to the fact that by iterative pruning, we let each client find its own partners (clients with similar labels) and leverage their personalized parameters to ameliorate their accuracy performance. In order to reveal how effective the proposed method is, we also compare the results versus two benchmarks, i.e., the \textit{Standalone}, and \textit{traditional FedAvg}. In the former, each client trains a model locally only by its own local data without federation. In the latter, all clients participate in the federation and the server averages the parameters traditionally by FedAvg. These results validate the effectiveness of our proposed algorithms for the objectives presented in the following Remarks:\\

\begin{figure*}[!h]
    \centering
    \begin{tabular}{ccc}
          \includegraphics[width=.24\pdfpagewidth]{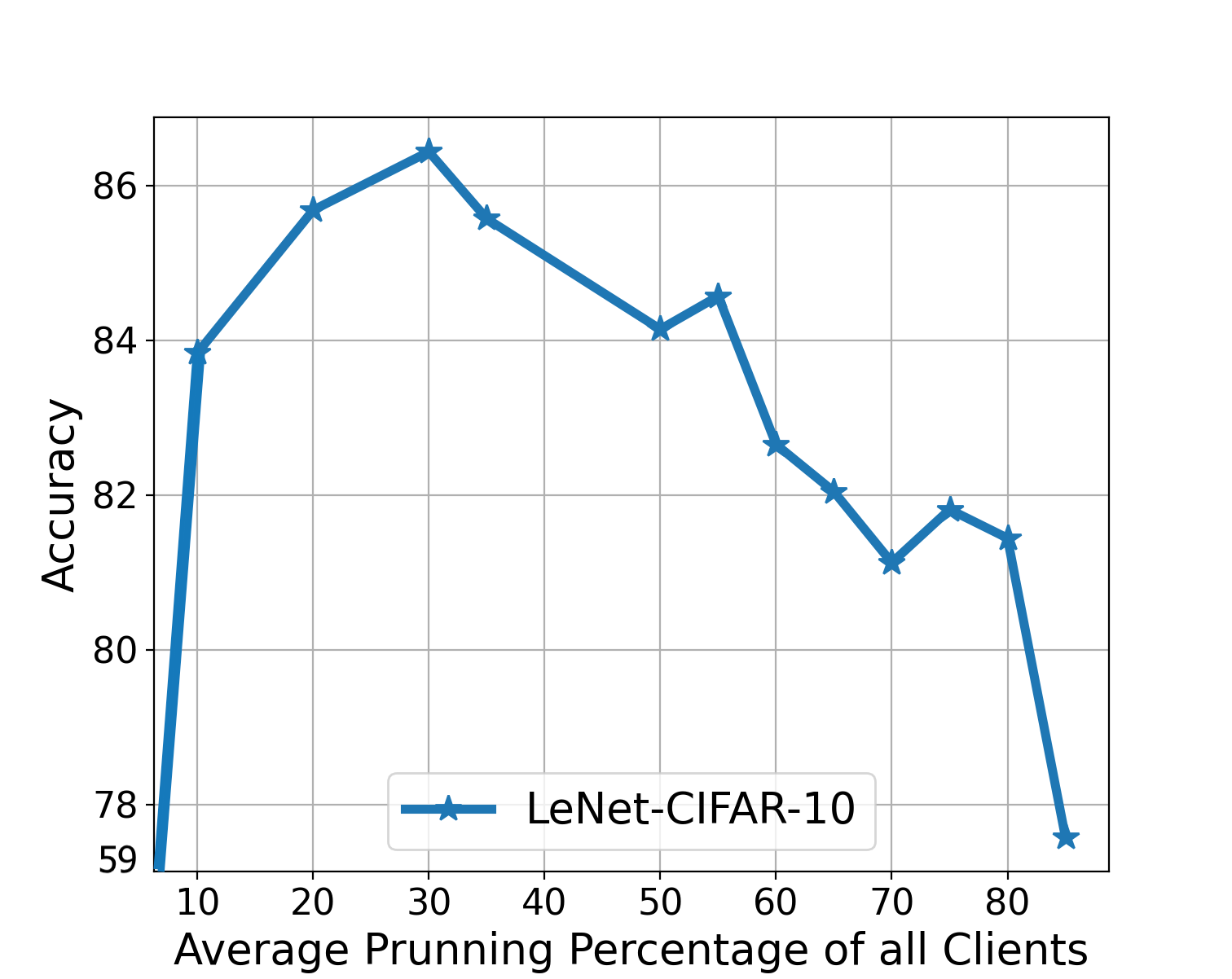} \hspace{-8mm} & 
          \includegraphics[width=.24\pdfpagewidth]{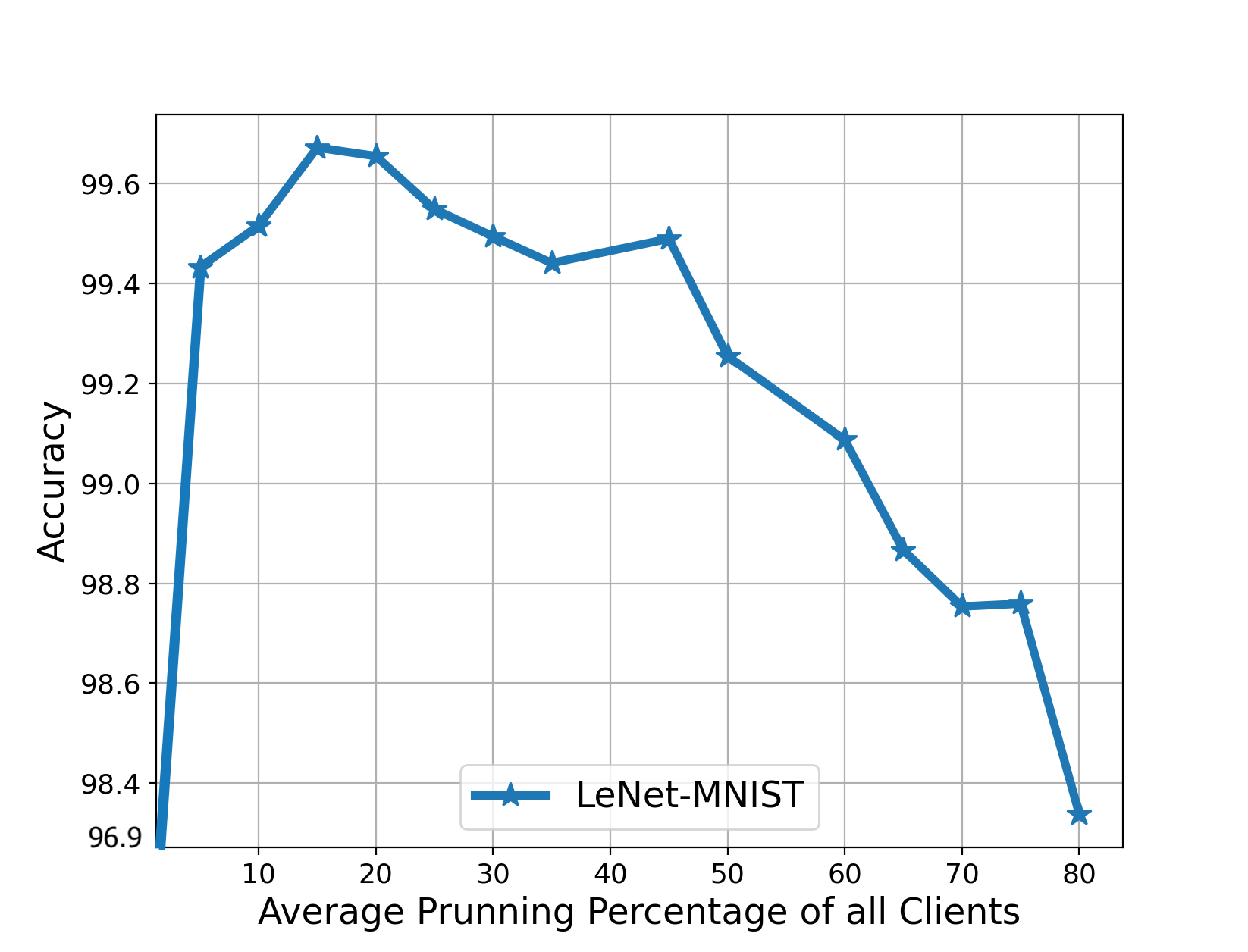} \hspace{-8mm}
          &
          \includegraphics[width=.24\pdfpagewidth]{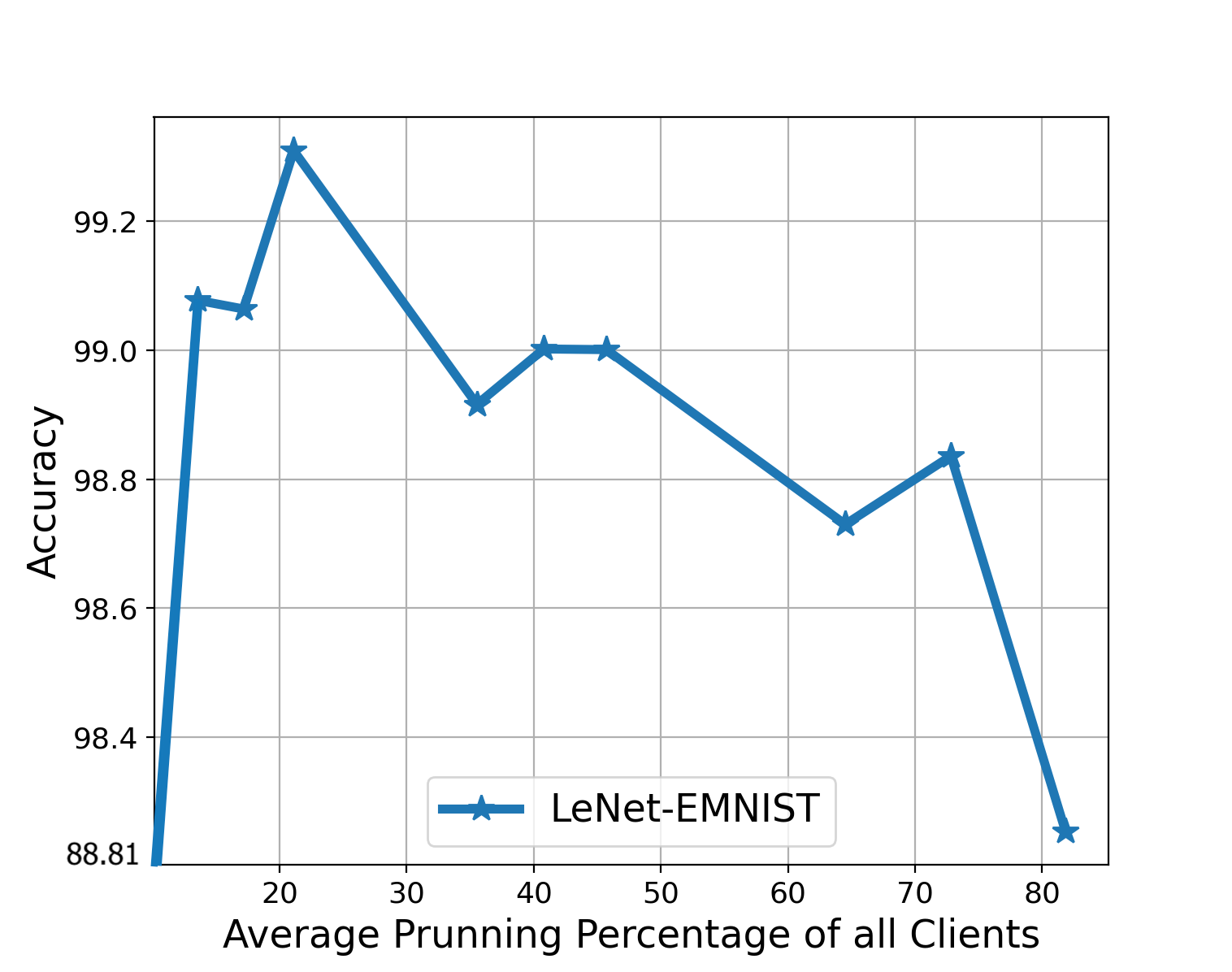}
    \end{tabular}
    \caption{Average test accuracy result of the proposed Sub-FedAvg (Un) versus various average pruning percentages over all clients, on LeNet-5 for CIFAR-10, MNIST and EMNIST benchmarks.}\label{fig2}
\end{figure*}

\begin{figure*}[!h]
    \centering
    \begin{tabular}{ccc}
          \includegraphics[width=.24\pdfpagewidth]{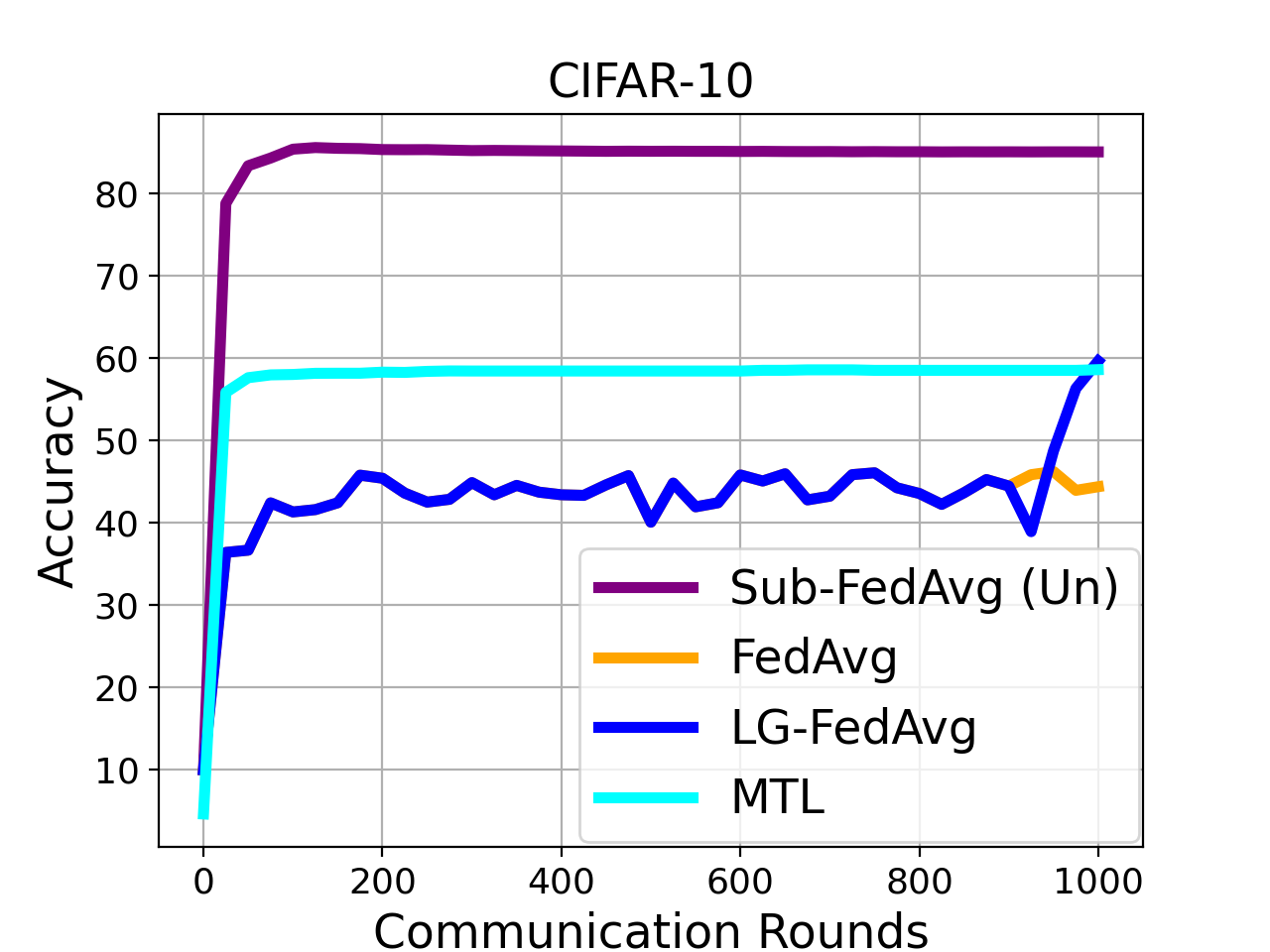} \hspace{-7mm} & 
          \includegraphics[width=.24\pdfpagewidth]{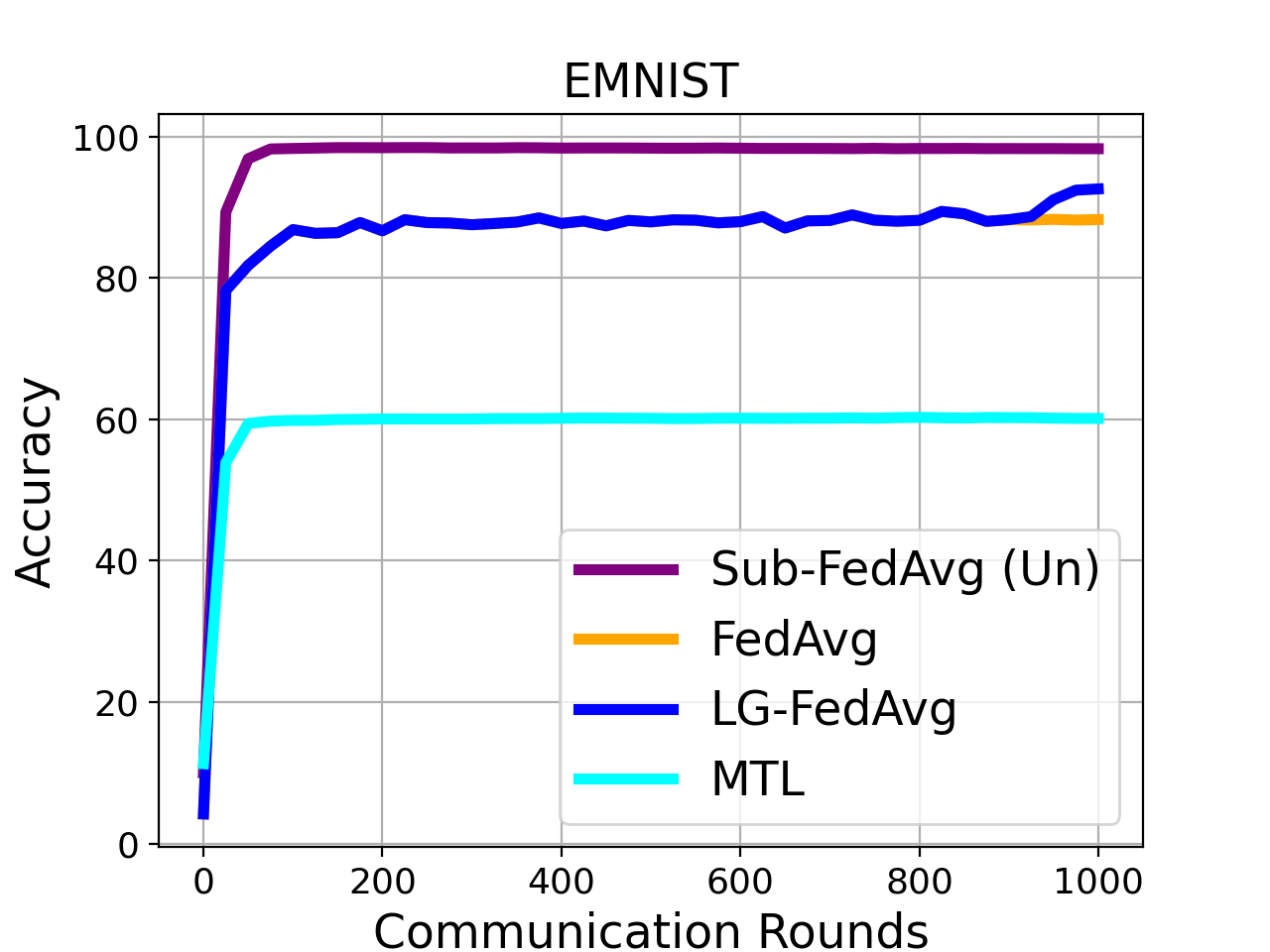}\hspace{-7mm} & 
           \includegraphics[width=.37\textwidth]{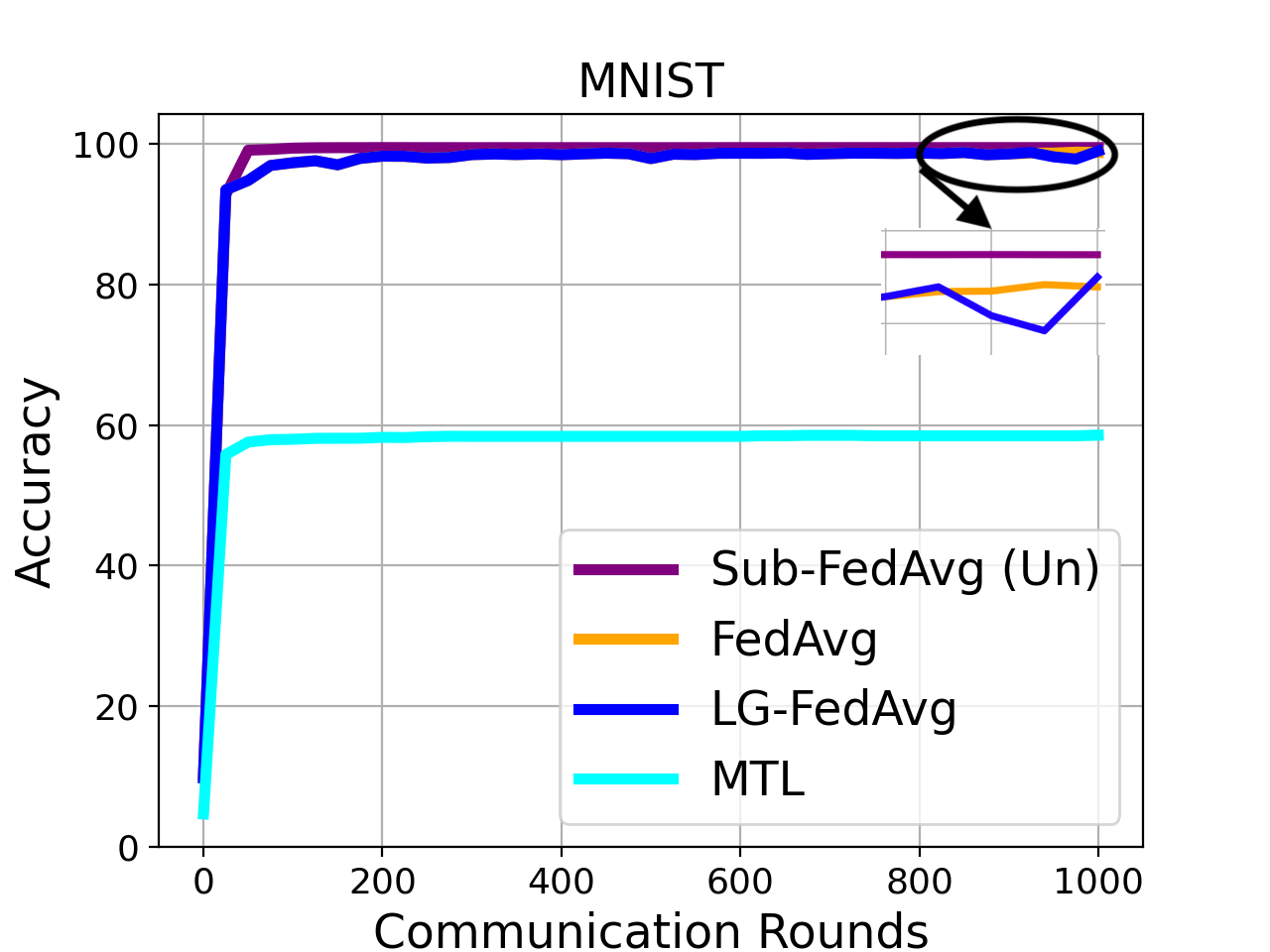}\\
    \end{tabular}
    \caption{Test accuracy vs. communication rounds for the CIFAR-10, EMNIST, and MNIST experiments under statistical heterogeneity. }\label{fig-commun}
\end{figure*}

\noindent \textbf{Remark-2:} In the low data regime, which is the case in most FL scenarios, where the clients are the edge devices, by structured/unstructured pruning (finding subnetworks) we get rid of the common parameters and keep the personalized ones to let the clients take the advantage of the featured data of the rest of clients with similar labels. This yields around $20\%$ accuracy improvements versus the \textit{Standalone} benchmark which further motivates federation. Compared to the traditional FedAvg benchmark, our results demonstrate around $30\%$ improvement. As is evident from the table, in the low data non-IID settings, the traditional FedAvg performs worse than the \textit{Standalone}, which means that it is not justified for clients to participate in federation under this setting. 

This phenomenon can be explained by the fact that depending upon the data (labels) of each client, 
by our pruning strategy, which prunes the common parameters and keep the personal ones, we let the clients to find their subnetwork. Through comprehensive experiments, we noticed that it is likely that the clients with label overlap share the same personal parameters. With that, we help all clients to find their corresponding partners in the federation and do the parameter averaging by Sub-FedAvg. With that in mind, the poor performance of the traditional FedAvg in non-IID environment can be understood.\\

\noindent \textbf{Remark-3:} Our proposed model also contributes in producing a more compressed model which achieves inference speedup compared to the original model. As can be seen from the table, our proposed model highly reduced the number of FLOPs up to $2.4 \times $ compared to the state-of-the-art. This is a crucial metric especially when the clients are the edge devices with computational limitations.

\subsubsection{Accuracy versus pruning percentage}
We have provided experiments on demonstrating how our algorithm can improve the accuracy of the clients. Fig.~\ref{fig-subnet} plots the test accuracy result of some sampled clients versus various pruning percentages when iteratively pruning by $5\%$-$10\%$ per iteration for CIFAR-10 on LeNet-5. Also, in Fig.~\ref{fig2} we sketched the average test accuracy over all clients versus various average pruning percentages over all clients. As expected, as the pruning target increases up to a certain level of sparsity, according to Remark-1 and Remark-2, the test accuracy rises as compared to the original network. According to our experiments almost $85\%$ of the clients have the same pattern of results. As expected, after some pruning rate (around $50\%$) we observe the accuracy degradation which is due to the fact that the by further pruning we are removing the personal parameters as well.

\begin{table*}[t]
\centering
\caption{Comparing the performance metrics of our results for the proposed algorithms, Sub-FedAvg (Hy), and Sub-FedAvg (Un) against the state-of-the-arts, i.e., FedAvg~(\cite{mcmahan2017communication}), MTL~(\cite{smith2017federated}), FedProx~(\cite{li2018federated}), and LG-FedAvg~(\cite{liang2020think}), on different datasets.}
\begin{tabular}{@{}cccccccc@{}}\toprule \toprule
\textbf{Setting} & Algorithm& Accuracy 
&\makecell{Pruned \% :\\ \%Filter+\%Parameters} 
& \makecell{Unstructured:\\ \% Parameters}
& \makecell{Communication\\ Cost}
& 
\\ \midrule \midrule
\textbf{LeNet-5}\\ \textbf{CIFAR-10} &Standalone &$84.44\%$ & 0 & 0 & 0 \\
&FedAvg &58.99\%  &0 & 0& 2.48 GB\\
&MTL & $49.87\%$& 0& 0& 16.12 GB\\
&LG-FedAvg & $76.28\%$& 0& 0& 2.27 GB \\
&Sub-FedAvg (Un)  & \textbf{86.01\%} & - & $30\%$ & 2.12 GB  \\
&Sub-FedAvg (Un)   & \textbf{84.44\%} & - & $50\%$ & 1.88 GB  \\
&Sub-FedAvg (Un)   & \textbf{83.6\%} & - & 70\% & 1.64 GB \\

&Sub-FedAvg (Hy)  & \textbf{83.21\%} & 49\% + 47.26\% = 50\% & 47.26\% & 1.89 GB  \\
&Sub-FedAvg (Hy)  & \textbf{82.86\%} & 47\% + 70\% = 73\% & 70\% & 1.62 GB \\
&Sub-FedAvg (Hy)  & \textbf{82.5\%} & 47\% + 90\% = 92\% & 90\% & 1.39 GB \\
\bottomrule
\textbf{LeNet-5}\\ \textbf{MNIST} &Standalone &94.25\% & 0 & 0 & 0  \\
&FedAvg & $96.9\%$  & 0 & 0& 524.16 MB  \\
&MTL & 99.74 & 0& 0& 3407.04 MB\\
&FedProx & $97.9\%$ & 0& 0& 1572.48 MB\\
&LG-FedAvg & $98.2\%$& 0& 0& 513.6 MB \\
&Sub-FedAvg (Un) &  $\textbf{99.43\%}$& -& $30\%$& 448 MB\\
&Sub-FedAvg (Un) &  $\textbf{99.28\%}$& -& 50\%& 397.21 MB\\
&Sub-FedAvg (Un) &  \textbf{99.35\%}& -& $70\%$& 346.43 MB\\
&Sub-FedAvg (Hy)  & \textbf{99.57\%}& $46\%$ + 48\%=50\%& 49\%& 383.39 MB\\
&Sub-FedAvg (Hy)  & \textbf{99.54\%}& $48\%$ + 70\% = 71\%& 70\%& 342.31 MB\\
&Sub-FedAvg (Hy)  & $\textbf{97.46\%}$& $40\% + 95\% = 90\%$& $95\%$ & 293.40 MB\\
\bottomrule
\textbf{LeNet-5}\\ \textbf{EMNIST}&Standalone & $98.59\%$ & 0 & 0 & 0 \\
&FedAvg & $88.81\%$ &0 & 0& 524.16 MB  \\
&MTL & 98.57& 0& 0& 3407.04 MB \\
&LG-FedAvg & $98.93\%$ & 0& 0& 513.6 MB \\
&Sub-FedAvg (Un) & $\textbf{99.11\%}$& -& $30\%$& 448 MB \\
&Sub-FedAvg (Un)  & $\textbf{99.16\%}$& - & $50\%$ & 397.21 MB\\
&Sub-FedAvg (Un)  & $\textbf{97.71\%}$& - &$ 70\%$ & 346.43 MB \\
&Sub-FedAvg (Hy)  & $\textbf{99.47\%}$ & $42\% + 42\% = 50\%$ & $42\%$ & 397.08 MB \\
&Sub-FedAvg (Hy)  & $\textbf{99.45\%}$ & $35\% + 69\% = 70\% $& 69\% & 344.26 MB   \\
&Sub-FedAvg (Hy)  &$\textbf{98.56\%}$ & $38\% + 93\% = 90\%$ & $93\%$ & 297.32 MB \\
\bottomrule

\textbf{LeNet-5}\\ \textbf{CIFAR-100}&Standalone & $80.56\%$ & 0 & 0 &0 \\
&FedAvg & $10.4\%$ &0 & 0& 2.78 GB \\
&MTL & $43.86\%$ & 0& 0&18 GB\\
&LG-FedAvg & $47.6\%$ & 0& 0&2.58 GB\\
&Sub-FedAvg (Un)   & $\textbf{85.5\% }$ & - & $30\%$ & 2.38 GB \\
&Sub-FedAvg (Un)  & $\textbf{83.40\% }$ & - & $50\% $& 2.11 GB  \\
&Sub-FedAvg (Un)  & $\textbf{83.74\% }$ & - & $70\%$ & 1.84 GB \\
&Sub-FedAvg (Hy) & $\textbf{82.16\% }$ & $48\% + 49\% = 50\%$  & $49\%$ & 2.12 GB  \\
&Sub-FedAvg (Hy)  & $\textbf{82.06\% }$  & $50\% + 69\% = 70\%$ & 69\% & 1.82 GB   \\
&Sub-FedAvg (Hy)  & $\textbf{80.80\% }$ & $50\% + 88\% = 88\%$ & $88\%$ & 1.56 GB \\
\bottomrule
\bottomrule
\end{tabular}
\label{tabel-1}
\end{table*}

\begin{table*}[t]
\centering
\caption{Comparing flop, and parameter reduction results for the proposed algorithms, Sub-FedAvg (Hy), and Sub-FedAvg (Un) against the state-of-the-arts on CIFAR-10,
MNIST, 
EMNIST, and
CIFAR-100 datasets.}
\vspace{2mm}
\begin{tabular}{@{}ccccc@{}}\toprule \toprule
Algorithm& Flop, Parameter Reduction&
\\ \midrule \midrule
Standalone  & $0\times$ \\
FedAvg~(\cite{mcmahan2017communication})  &$0\times$\\
MTL~(\cite{smith2017federated})& $0\times$\\
LG-FedAvg~(\cite{liang2020think})&  $0\times$\\
Sub-FedAvg (Un), $p_{us}=30$  & $0 \times ,  0.3 \times$ \\
Sub-FedAvg (Un), $p_{us}=50$   & $0 \times ,  0.5 \times$  \\
Sub-FedAvg (Un),  $p_{us}=70$    & $0 \times ,  0.7 \times$  \\
Sub-FedAvg (Hy), $p_{s}=50$   & $2.4 \times,  0.5 \times$  \\
Sub-FedAvg (Hy), $p_{s}=70$  & $ 2.4 \times,  0.7 \times$  \\
Sub-FedAvg (Hy), $p_{s}=90$  & $ 2.4 \times, 0.9 \times$  \\
\bottomrule
\bottomrule
\end{tabular}
\label{tabel-1}
\end{table*}

\subsubsection{Communication Efficiency}
In FL, communication cost is an important bottleneck since the edge devices are typically constrained by an upload bandwidth of $1$ MB/s or less.
The results in Table~\ref{tabel-1} clearly demonstrates that the proposed Sub-FedAvg (Hy) and Sub-FedAvg (Un) methods provide significant performance improvement in terms of communication cost across prior benchmarks reported in the table. It is noteworthy that we obtained the communication cost for all of the baselines as $Cost=R\times B \times |W| \times 2$, where, $R$ stands for the total number of communication rounds; $B$ denotes the number of bits based on single-precision floating point (32 bit for floating numbers and 1 bit for integers $0$ and $1$); and $|W|$ represents the total number of model parameters that are exchanged between each client and the server in each round. Unlike prior works~(\cite{shokri2015privacy}) where the communication cost is addressed by only sharing a subset of the parameters during each round of communication by sacrificing the performance, our proposed model can produce a dramatic decrease in communication cost without sacrificing the accuracy due to the following facts: 

Firstly, our model prunes common parameters without impacting the accuracy of clients and keeps the personalized ones as mentioned in Remark-2. Hence, the server collects an efficient aggregation of the clients' updates by only collecting the kept ones.
Secondly, by the proposed Sub-FedAvg (Hy) and Sub-FedAvg (Un) we end up with getting a compressed version of the neural networks which work efficiently and converge faster. In our experiments, for the Sub-FedAvg (Un) method we got the reported results for CIFAR-10/100, MNIST, and EMNIST datasets by training the neural networks only for 500, 300, and 300 communication rounds, respectively. This is while other baselines typically perform around $1000$ rounds of communication to converge to their desired accuracy performance. This fast convergence along with the mentioned fact, contributed in the reduction of required communication rounds by up to $10 \times$ compared to the prior works without accuracy degradation. Fig~\ref{fig-commun} shows the comparison of our method and the state-of-the-art in terms of accuracy versus the number of communication rounds. We achieved around $99.5\%$, $99.3\%$, and $86\%$ test accuracy for MNIST, EMNIST, and CIFAR-10 datasets, respectively, after around $100$ rounds, while the traditional FedAvg, LG-FedAvg, and MTL achieve a lower test accuracy after $100$-$800$ communication rounds. We observe that Sub-FedAvg (Un) achieves a significant reduction in required communication round by $2$-$10$ times in all cases. For CIFAR-10, FedAvg requires the highest number of communication rounds to obtain the target accuracy, which is about $3$ times more than that of our method.


\subsubsection{Flop Reduction}
To further highlight our proposed method's efficiency we delve into the percentage of parameter saving and FLOP saving. In speedup analysis, operations such as batch normalization (BN) and pooling are ignorable comparing to convolution operations. As such, we count the FLOPs of convolution operations for computation complexity analysis, which is common trend in prior works~(\cite{liu2017learning}).
In our experiments, for example, for LeNet-5, $50\%$ of channels pruned which results in around $50\%$ FLOP reduction while the parameter saving is around $38\%$. This is due to the fact that for example in LeNet-5 only 11 (out of 22) channels from all the computation-intensive convolutional layers are pruned, while 24k parameters (out of 49k) from the parameter-intensive fully-connected layers are pruned. The results show that our methods not only work efficient with much fewer filters and channels, and save significant number of FLOPs and accelerate the inference but also leverage it for accuracy improvement on CIFAR-10/100, MNIST, and EMNIST datasets. 


%% file: sections/Conclusion.tex
\section{Conclusion}

A new framework for personalized federated learning with Non-IID data distributions was proposed. The method works by iteratively pruning the parameters and channels of the neural networks which results in removing the commonly shared parameters of clients' model and keeping the personalized ones. In contrast to other approaches, the proposed framework is also efficient in terms of communication cost and FLOP count. We found that our method outperforms the state-of-the-art algorithms on CIFAR-10/100, MNIST, and EMNIST benchmarks.

\section{Acknowledgement}
This research is based upon work supported in parts by the
National Science Foundation under NSF IIS-1956339. The views, findings,
opinions, and conclusions contained herein are those of the authors and should not be interpreted as necessarily representing the official policies or endorsements, either expressed or implied, of the NSF or the U.S. Government.